\pdfoutput=1
\documentclass[letterpaper]{article} 
\usepackage{aaai2026}  
\usepackage{times}  
\usepackage{helvet}  
\usepackage{courier}  
\usepackage[hyphens]{url}  
\usepackage{graphicx} 
\usepackage{natbib}  
\usepackage{caption} 
\usepackage{algorithm}
\usepackage{algorithmic}

\usepackage{newfloat}
\usepackage{listings}
\DeclareCaptionStyle{ruled}{labelfont=normalfont,labelsep=colon,strut=off} 
\floatstyle{ruled}
\newfloat{listing}{tb}{lst}{}
\floatname{listing}{Listing}
\usepackage{amsmath}
\usepackage{pifont}
\usepackage{booktabs}
\usepackage{amsfonts}
\usepackage{hyperref}
\usepackage{cleveref}
\crefname{figure}{Fig.}{Figs.}
\crefname{table}{Table.}{Tables.}
\crefname{section}{Section}{Secs.}

\title{D-FCGS: Feedforward Compression of Dynamic Gaussian Splatting for Free-Viewpoint Videos}
\author {
    Wenkang Zhang\textsuperscript{\rm 1, \rm 2},
    Yan Zhao\textsuperscript{\rm 1},
    Qiang Wang\textsuperscript{\rm 3},
    Zhixin Xu\textsuperscript{\rm 1}, 
    Li Song\textsuperscript{\rm 1}, 
    Zhengxue Cheng\textsuperscript{\rm 1}\thanks{Corresponding Author}
}
\affiliations {
    \textsuperscript{\rm 1}Shanghai Jiao Tong University\\
    \textsuperscript{\rm 2}Zhiyuan College, Shanghai Jiao Tong University\\
    \textsuperscript{\rm 3}Visionstar Information Technology (Shanghai) Co., Ltd\\
    \{conquer.wkzhang, zhaoyanzy, zhixin.xu, song\_li, zxcheng\}@sjtu.edu.cn, wq@sightp.com
}

\usepackage{bibentry}

\begin{document}

\maketitle

\begin{abstract}
Free-Viewpoint Video (FVV) enables immersive 3D experiences, but efficient compression of dynamic 3D representation remains a major challenge.
Existing dynamic 3D Gaussian Splatting methods couple reconstruction with optimization-dependent compression and customized motion formats, limiting generalization and standardization. 
To address this, we propose D-FCGS, a novel \textbf{F}eedforward \textbf{C}ompression framework for \textbf{D}ynamic \textbf{G}aussian \textbf{S}platting. Key innovations include: (1) a standardized Group-of-Frames (GoF) structure with I-P coding, leveraging sparse control points to extract inter-frame motion tensors; (2) a dual prior-aware entropy model that fuses hyperprior and spatial-temporal priors for accurate rate estimation; (3) a control-point-guided motion compensation mechanism and refinement network to enhance view-consistent fidelity. 
Trained on Gaussian frames derived from multi-view videos, D-FCGS generalizes across diverse scenes in a zero-shot fashion. Experiments show that it matches the rate-distortion performance of optimization-based methods, achieving over 17$\times$ compression compared to the baseline while preserving visual quality across viewpoints. This work advances feedforward compression of dynamic 3DGS, facilitating scalable FVV transmission and storage for immersive applications.
\end{abstract}

\begin{links}
    \link{Code}{https://github.com/Mr-Zwkid/D-FCGS}
\end{links}

\section{Introduction}
Our world is inherently dynamic, with 3D scenes evolving over time and observable from arbitrary viewpoints. Capturing and representing such 4D dynamics has long been a fundamental challenge in computer vision and graphics. Free-viewpoint video (FVV), which enables immersive 6-DoF experiences, has emerged as a promising solution with applications in virtual reality, telepresence, and remote education. However, realizing practical FVV systems demands efficient solutions for reconstruction, compression, transmission, and rendering. In this work, we focus on the critical problem of compression for dynamic 3D scenes.

Recent advances of 3D Gaussian Splatting (3DGS) \cite{kerbl20233d} have revolutionized 3D scene representation, offering unparalleled rendering quality and real-time performance. Extending to 4D, dynamic forms of 3DGS \cite{yang2024deformable, wu20244d, li2024spacetime, yang2023real, sun20243dgstream} gradually garner attention. Analogous to the temporal expansion of images facilitated by videos, frame-by-frame 3D Gaussians can naturally serve as a temporal expansion of 3DGS, forming the basis for FVV. Based on this per-frame idea, 3DGStream \cite{sun20243dgstream} pioneers dynamic scene reconstruction via on-the-fly training, while subsequent works improve compression efficiency through techniques like rate-aware training \cite{hu20254dgcrateaware4dgaussian}, vector-quantized residuals \cite{girish2024queen}, and static-dynamic decomposition \cite{wu2025swift4d}.

\begin{figure}[t]
  \centering
   \includegraphics[width=1\linewidth]{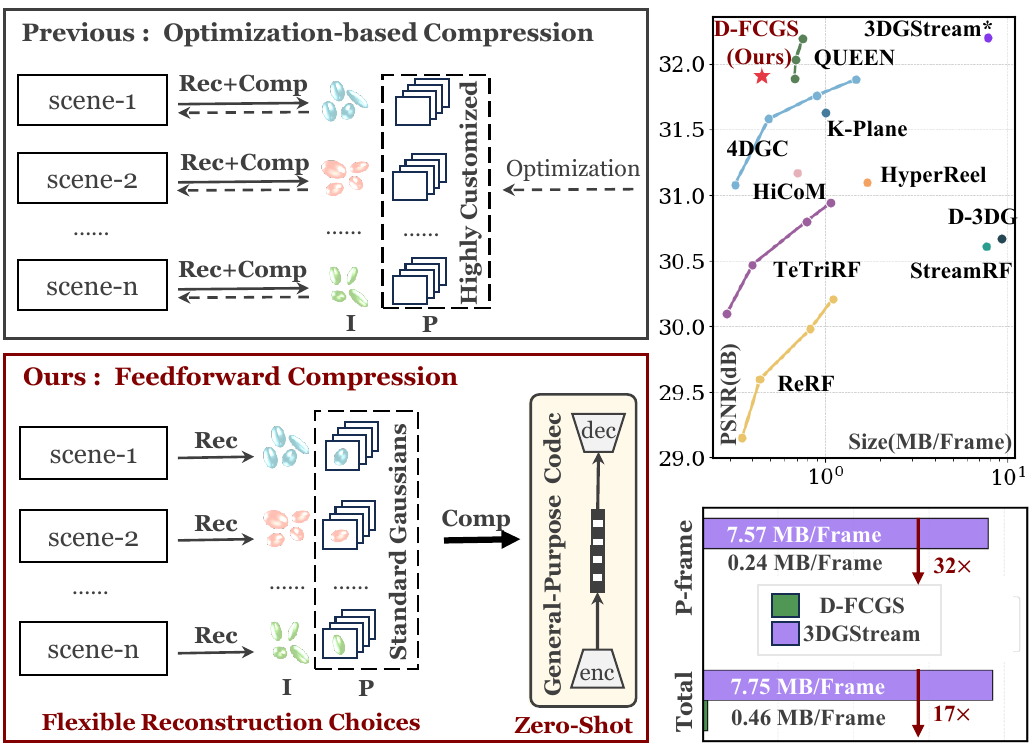}

   \caption{ \textbf{Left:} Illustration of differences between previous optimization-based methods and our D-FCGS. \textbf{Right:} R-D curve and storage size comparison on N3V dataset.
    }
   \label{fig:teaser}
\end{figure}

Despite these advances, existing methods remain constrained by their coupled optimization of reconstruction and compression. By representing frame-wise Gaussian motions through specialized formats (e.g., neural networks), these approaches require per-scene optimization and meticulous hyperparameter tuning. This optimization-dependent paradigm creates two critical barriers to practical deployment: (1) it severely limits generalization to unseen scenes; (2) it hinders the development of standardized compression schemes that could enable widespread adoption of FVV. 

To address these challenges, we propose \textbf{D-FCGS}, a novel  feedforward compression framework for dynamic Gaussian Splatting. Our key insight is that temporal coherence in Gaussian point clouds can be efficiently modeled via the Group-of-Frames (GoF) representation, where inter-frame motions can be compressed in a feedforward and scene-agnostic manner. Specifically, we adopt the standard GS format as input and leverage sparse control points to efficiently extract motion tensors, optimal for both compression efficiency and computational performance. These motion tensors are then processed through our feedforward motion compression pipeline, which incorporates a tailored dual prior-aware entropy model to enhance probability estimation accuracy. Following decompression, the sparse motions are propagated across the entire Gaussian frame under the guidance of the control points. Finally, lightweight color refinement is applied to improve view-consistent fidelity.

Our D-FCGS is trained end-to-end on frame-wise GS sequences constructed from both real-world and synthetic multi-view video datasets. Once trained, it serves as \textbf{a general-purpose inter-frame compression codec}, requiring no scene-specific optimization or access to multi-view images during zero-shot inference. Extensive experiments show that D-FCGS exhibits strong generalization across diverse dynamic scenes and achieves state-of-the-art rate-distortion performance, surpassing 17$\times$ compression over 3DGStream while maintaining comparable fidelity.

Our contributions can be summarized as follows: 
\begin{itemize}
\item We present D-FCGS, a novel feedforward compression framework for dynamic 3DGS that enables zero-shot inter-frame coding of Gaussian sequences.
\item We propose a sparse motion representation with control points for I-P coding, coupled with a dual prior-aware entropy model for efficient compression, and develop a decoder with motion compensation and refinement for enhanced fidelity.
\item Experiments across various scenes show the effectiveness and robustness of D-FCGS, achieving over 17$\times$ compression over 3DGStream while preserving fidelity, outperforming most optimization-based methods.
\end{itemize}

\section{Related Work}

\subsection{3D Gaussian Splatting Compression} 

High storage demands of 3DGS have motivated compression efforts. Optimization-based compression methods include value-based and structure-based ones. The former involve the use of pruning \cite{girish2024eagles, ali2024trimming}, masking \cite{lee2024compact, wang2024end}, vector quantization \cite{niedermayr2024compressed, navaneet2023compact3d} and distillation \cite{fan2024lightgaussian} to reduce insignificant Gaussians or redundancy of Gaussian parameters. The latter utilize structural modeling such as anchors \cite{lu2024scaffold}, tri-planes \cite{lee2025compression} and 2D grids \cite{morgenstern2024compact} to address the sparsity and unorganized nature of Gaussian Splatting. When combined with learned entropy models \cite{balle2016end, cheng2020learned}, they achieve excellent rate-distortion performance \cite{chen2024hac, chen2025hac++, zhan2025cat}.

However, these optimization-based methods require per-scene fine-tuning, limiting their generalization. Recent works have introduced feedforward pipelines \cite{chen2024fast,  huang2025hierarchical, yang2025hybridgs} that decouples reconstruction and compression. These pre-trained codecs can directly compress arbitrary Gaussian clouds without multi-view supervision, offering greater practicality. We aim to extend this paradigm to dynamic GS compression.

\begin{figure*}[ht]
  \centering
   \includegraphics[width=1\linewidth]{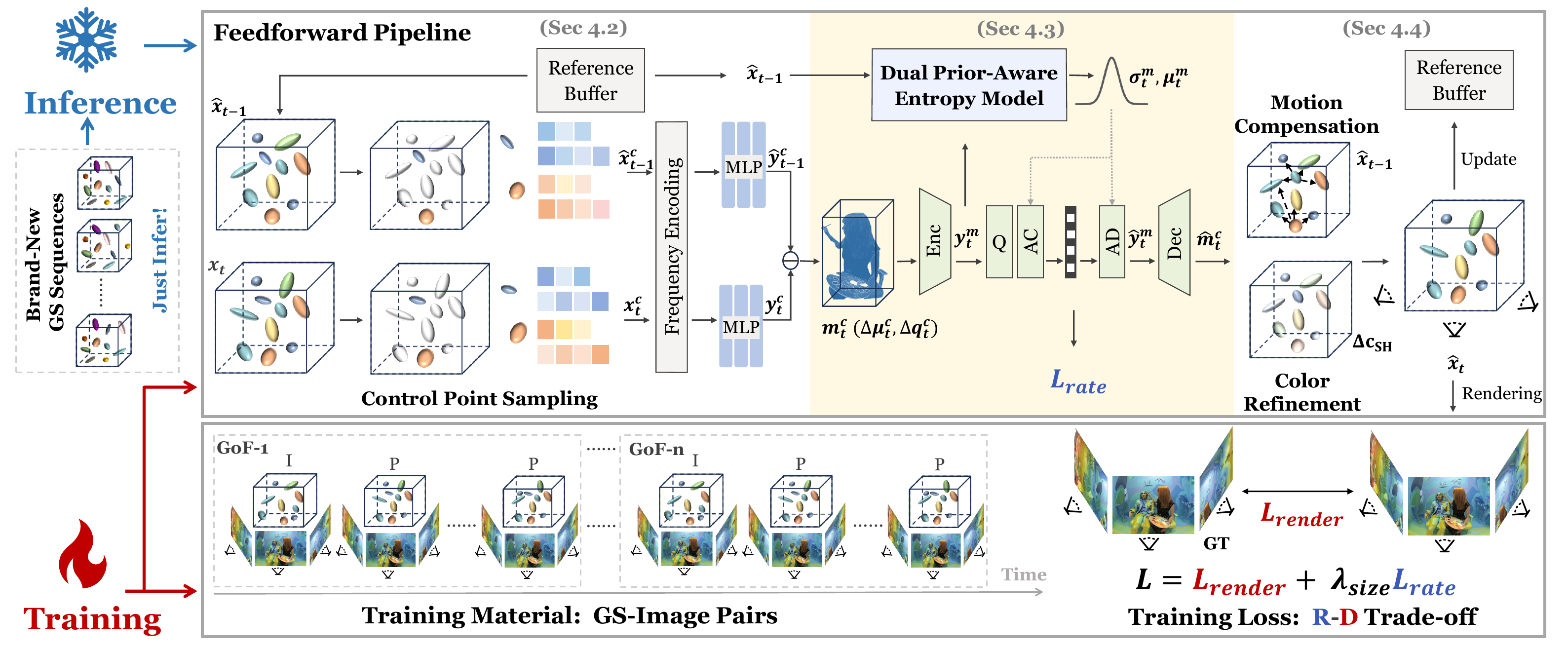}

   \caption{ \textbf{Overview of D-FCGS framework.} Our feedforward pipeline processes sequential Gaussian frames in GoF through three stages: (1) sparse motion extraction (\cref{sec:motion_extraction}), (2) feedforward motion compression (\cref{sec:motion_compression}), and (3) motion compensation and refinement (\cref{sec:frame_recon}). Once trained with rate-distortion loss, D-FCGS can infer on brand-new GS sequences in a zero-shot manner.}

   \label{fig:overview}
\end{figure*} 

\subsection{Dynamic GS and its Compression}

Dynamic GS methods can be categorized by their motion representation approaches. Implicit/explicit motion fields \cite{yang2024deformable, wu20244d, li2024spacetime, lin2024gaussian} and 4D Gaussian formulations \cite{duan20244d, lee2024fully} often face challenges in real-time streaming, variable resolutions, or long durations. In contrast, per-frame approaches \cite{luiten2024dynamic, sun20243dgstream, wang2024v, yan2025instant} demonstrate superior practicality through incremental frame updates, making them ideal for free-viewpoint video streaming. Our work focuses on this line of \textit{Per-Frame Gaussian} methods.

For compression of \textit{Per-Frame Gaussian}, HiCoM \cite{gao2024hicom} employs hierarchical grid-wise motion representation and QUEEN \cite{girish2024queen} leverages a latent-decoder for quantization of attribute residuals. 4DGC \cite{hu20254dgcrateaware4dgaussian} instills rate-aware training into frame reconstruction, achieving decent compression. While effective, they fundamentally couple reconstruction with compression, requiring scene-specific tuning. Our work breaks this constraint by introducing a novel optimization-free compression framework for dynamic GS, enabling zero-shot compression to arbitrary 4D scenes.

\section{Preliminary}
Our feedforward compression method takes frame-wise standard 3DGS as input, denoted as Gaussian frames, and each frame is pre-optimized from multi-view images at the respective timestamp.  Within a given Gaussian frame, each Gaussian primitive is characterized by: (1) \textbf{geometry parameters} including position $\boldsymbol{\mu} \in \mathbb{R}^3$ and covariance matrix $\boldsymbol{\Sigma} \in \mathbb{R}^{3\times3}$; (2) \textbf{appearance parameters} including opacity $o \in \mathbb{R}^1$ and SH-based color $c_{SH} \in \mathbb{R}^{48}$. The covariance matrix can be further represented as $\boldsymbol{\Sigma } = \boldsymbol{R}\boldsymbol{S}\boldsymbol{S}^T\boldsymbol{R}^T$, where $\boldsymbol{R} \in \mathbb{R}^{3\times3}$ is the rotation matrix parameterized by the quaternion $\boldsymbol{q} \in \mathbb{R}^4$,  and the scale matrix $\boldsymbol{S} \in \mathbb{R}^{3\times3}$ is a diagonal matrix with elements $\boldsymbol{s} \in \mathbb{R}^3$. The geometry of a Gaussian primitive can be formulated as:
\begin{equation}
  G(\boldsymbol{x}) = e^{-\frac{1}{2}(\boldsymbol{x} - \boldsymbol{\mu})^T \boldsymbol{\Sigma}^{-1} (\boldsymbol{x} - \boldsymbol{\mu})}, 
\end{equation}
where $\boldsymbol{x} \in \mathbb{R}^3$ is any random 3D location within the scene. 

Given a viewpoint, 3D Gaussians are projected into a 2D plane, and the color of a pixel $C \in \mathbb{R}^3$ is derived by alpha-blending of overlapping 2D Gaussians:
\begin{equation}
  C = \sum_i c_i \alpha_i \prod_{j=1}^{i-1} (1-\alpha_j), 
\end{equation}
where $c_i \in \mathbb{R}^3$ is the view-dependent color calculated from SH-based color $c_{SH}$, and $\alpha_i \in \mathbb{R}^1$ is the blending weight derived from opacity $o$.

With the differentiable rasterizer, training can be supervised by 2D images from varied views in an end-to-end way. The rendering loss of vanilla 3DGS is:
\begin{equation}
  L_{\text{render}} = \lambda L_{\text{D-SSIM}} + (1-\lambda) L_1.
\end{equation}

\section{Method}
\subsection{Overview}

As shown in \cref{fig:overview}, our D-FCGS framework comprises three key components: (1) sparse motion extraction (\cref{sec:motion_extraction}), (2) feedforward motion compression (\cref{sec:motion_compression}) and (3) motion compensation and refinement (\cref{sec:frame_recon}). Training and inference procedures will be mentioned in \cref{sec:training}.

Given two adjacent Gaussian frames $\boldsymbol{x_t}$ and $\boldsymbol{\hat{x}_{t-1}}$, we first sample sparse control points from dense Gaussians and extract motion features. These motion tensors are compressed using our dual prior-aware entropy model for efficient rate estimation. During decoding, we reconstruct $\boldsymbol{\hat{x}_t}$ via control-point-guided motion compensation and color refinement, while storing  it into the buffer for future reference.

\subsection{Sparse Motion Extraction via Control Points}
\label{sec:motion_extraction}
Building on observations that most Gaussians exhibit local motion coherence, we propose an efficient sparse representation. Inspired by previous optimization-based 4D reconstruction methods \cite{he2024s4d, yan2025instant, huang2024sc}, we employ Farthest Point Sampling (FPS) to select $N^c = \frac{N}{M}$ control points from the $N$ Gaussian primitives, and derive corresponding geometry parameters, formulated as:
\begin{equation}
    \boldsymbol{\mu^c} = FPS(\{\mu_i\}_{i\in N}, \frac{N}{M}),
\end{equation}
\begin{equation}
    \boldsymbol{x^{c}} = Index(\boldsymbol{x}, \boldsymbol{\mu^c}),
\end{equation}
where $M$ denotes the downscale factor, and $\boldsymbol{x^{c}}$  ($\boldsymbol{x^c} = \{\boldsymbol{\mu^c}, \boldsymbol{q^c}\}$) represent geometry parameters of control points. This process efficiently reduces storage demand and processing costs, while preserving motion characteristics.

After control point sampling, we derive the native sparse geometry attributes of the current frame and reference Gaussian frame, denoted as $\boldsymbol{x^c_t}$ and $\boldsymbol{\hat{x}^c_{t-1}}$. We encode these attributes using frequency encoding \cite{mildenhall2021nerf} and MLP projection:
\begin{equation}
    \boldsymbol{y^c_{t}} = MLP(FreqEnc\{\boldsymbol{x^c_t}\}),
\end{equation}
\begin{equation}
    \boldsymbol{\hat{y}^c_{t-1}} = MLP(FreqEnc\{\boldsymbol{\hat{x}^c_{t-1}}\}).
\end{equation}

Motion tensors are then derived in the feature domain: 
\begin{equation}
    \boldsymbol{m^c_{t}}   = Converter (\boldsymbol{y^c_{t}} - \boldsymbol{\hat{y}^c_{t-1})},
\end{equation}
where $\boldsymbol{m^c_{t}}$ denotes the motion tensors at time $t$, which keeps the same dimension as $\boldsymbol{x^c_t}$ and $\boldsymbol{\hat{x}^c_{t-1}}$. $Converter$ is realized by MLP at the feature level.

\subsection{Feedforward Motion Compression}
\label{sec:motion_compression}

\paragraph{\textbf{End-to-End Motion Compression.}}
The obtained sparse motion tensors $\boldsymbol{m^c_{t}}$ are then fed into our end-to-end compression module. This process begins with data encoding, followed by differentiable quantization simulated by additive uniform noise \cite{balle2016end} to enable gradient backpropagation :
\begin{equation}
    \begin{aligned}
    \boldsymbol{\hat{y}^m_t} =Q(\boldsymbol{y^m_t}) 
    & = \boldsymbol{y^m_t} + \mathcal{U}(-\frac{q'}{2}, \frac{q'}{2}),  \text{ for training} \\
    & = Round(\frac{\boldsymbol{y^m_t}}{q'}) \cdot q', \text{ for testing} \\
    \end{aligned}
\end{equation}
where $q'\in \mathbb{R}^1$ is the quantization step size, and $\boldsymbol{y^m_t}$, $\boldsymbol{\hat{y}^m_t}$ denote the encoded latent motion before and after quantization, respectively. Next to that, arithmetic coding (AC) converts the quantized data into a compact bitstream for efficient transmission and storage. On the decoder side, the bitstream is decompressed back into motion tensors using arithmetic decoding (AD) for frame reconstruction.

\paragraph{\textbf{Dual Prior-Aware Entropy Model.}}
According to Shannon’s theory \cite{shannon1948mathematical}, the cross-entropy between the estimated and true latent distributions provides a tight lower bound on the achievable bitrate:
\begin{equation}
    R(\boldsymbol{\hat{y}^m_t}) \ge \mathbb{E}_{\boldsymbol{\hat{y}^m_t} \sim q_{\boldsymbol{\hat{y}^m_t}}}[-\log _2p_{\boldsymbol{\hat{y}^m_t}}(\boldsymbol{\hat{y}^m_t})],
\end{equation}
where $p_{\boldsymbol{\hat{y}^m_t}}$ and $q_{\boldsymbol{\hat{y}^m_t}}$ are respectively estimated and true probability mass functions (PMFs) of the quantized latent codes $\boldsymbol{\hat{y}^m_t}$. Since arithmetic coding can achieve a bitrate close to this bound, our goal is to devise an entropy model that accurately estimates $p_{\boldsymbol{\hat{y}^m_t}}$.

\begin{figure}
    \centering
    \includegraphics[width=1\linewidth]{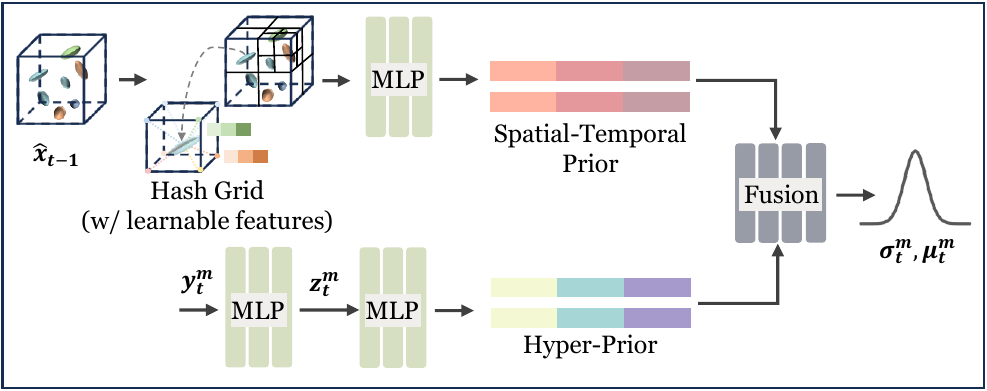}
    \caption{\textbf{Illustration of the dual prior-aware entropy model.} Spatial-temporal context priors extracted via multi-scale hashgrids and hyperpriors generated through the factorized model are fused by a lightweight fusion network for bitrate estimation.}

    \label{fig:entropy}
\end{figure}

\cref{fig:entropy} shows the proposed dual prior-aware entropy model that combines hyperprior and spatial-temporal context prior for precise distribution estimation. 
Following \cite{balle2018variational}, we use a factorized model to learn the hyperprior and estimate its PMF $p(\boldsymbol{\hat{z}^m_t})$, which is common in deep image and video compression. However, for GS-based FVV, the latent codes also exhibit strong 3D-spatial and temporal correlations. 
Assuming adjacent Gaussian frames share similar features, we strive to extract spatial-temporal priors from the reference frame $\boldsymbol{\hat{x}_{t-1}}$ through a multi-resolution hash grid encoding scheme. Specifically, at each grid intersection, we store learnable feature vectors that capture position-specific information. For each Gaussian positioned at $\boldsymbol{\mu_t}$, we retrieve multi-scale features by performing tri-linear interpolation across different levels $G^l$ of the voxel grid, where $l$ indicates the resolution level. The interpolated features from all levels are then concatenated and processed by a lightweight MLP to generate comprehensive positional contexts:
\begin{equation}
    context = MLP( \bigcup_{l=1}^L Interp(\boldsymbol{\mu_t}, G^l)),
\end{equation}
where $Interp(\cdot)$ denotes the grid interpolation operation. These positional contexts are subsequently combined with the transformed appearance parameters as the final spatial-temporal priors. The prior fusion network is then to integrate these spatial-temporal priors with the hyperpriors, estimating the mean $\boldsymbol{\mu^m_t}$ and scale $\boldsymbol{\sigma^m_t}$ of the latent code distribution (assumed as normal distribution). The probability mass for each quantized latent value $\hat{y}^m_{t,i}$ is computed by integrating the PMF over the quantization bin:
\begin{equation}
    p(\hat{y}^m_{t, i}) = \int_{\hat{y}^m_{t, i} - \frac{q'}{2}} ^ {\hat{y}^m_{t, i} + \frac{q'}{2}} \mathcal{N}(y|\mu^m_{t, i}, \sigma^m_{t, i}) dy,
\end{equation}
where $i$ corresponds to the index of a certain control point. The rate loss combines contributions from both the motion and hyper latent distributions:
\begin{equation}
    L_{\text{rate}} = \frac{1}{N^c} \sum_{i=1}^{N^c} (-\log_2( p(\hat{y}^m_{t, i}))  -\log_2(p(\hat{z}^m_{t, i}))),
\end{equation}
where $N^c = \frac{N}{M}$ is the total number of control points.

\subsection{Motion Compensation and Refinement}
\label{sec:frame_recon}

\paragraph{\textbf{Control Point Guided Motion Compensation.}}
The decoded control point motions $\boldsymbol{\hat{m}^c_t}$ are propagated to the entire Gaussian set in a distance-aware compensation way. For the $i^{th}$ control point, we first identify its $K$-nearest Gaussians $\mathcal{K}(i)$ via KNN search. The motion vectors are then distributed to neighboring Gaussians using an exponentially decaying weight function based on spatial distance. The closer the distance is, the more influence we consider the control point can exert on the motion of this neighboring point. The motion that $i^{th}$ control point assigns to its $j^{th}$ neighbor can be written as:
\begin{equation}
m_{i,j} = \frac{e^{-d_{i, j}} m_i}{\sum_{k\in \mathcal{K}(i)}e^{-d_{i, k}}},
\end{equation}
where $d_{i, j}$ represents the Euclidean distance between the $i^{th}$ control point and its $j^{th}$ neighbor. The aggregated motion vectors are finally applied to adjust the geometry parameters of the reference frame via addition. 

This approach provides two key benefits: (1) enhanced motion estimation accuracy through localized correlation modeling that suppresses error propagation, and (2) inherent parallelizability due to the independent processing of control points, ensuring computational efficiency. More discussions on frame matching during motion compensation can be seen in the supplementary.

\paragraph{\textbf{Color Refinement.}}
For image and video compression, post-compression refinement plays a critical role in mitigating visual artifacts such as color banding and blurring. 
In our GS-based framework, we specifically target refinement at SH coefficients $c_{SH}$ while preserving sensitive geometry and opacity parameters \cite{chen2024fast, girish2024queen, papantonakis2024reducing}.
The spatial-temporal priors from the entropy model module are repurposed to predict color residuals $\Delta c_{SH}$, which are dynamically added to $c_{SH}$ during decoding without additional storage. This \textbf{on-the-fly} refinement is fully differentiable, allowing gradient backpropagation through the entropy model for joint rate-distortion optimization.

\subsection{Training and Inference Pipeline of D-FCGS}
\label{sec:training}

\paragraph{\textbf{Training Process and Loss.}}
Following established practices \cite{wang2023neural, zheng2024hpc, zheng2024jointrf}, we adopt a Group-of-Frames (GoF) paradigm with the structure:
\begin{equation}
\scalebox{0.8}{$\scriptstyle
\underbrace{(I-P\!-\!\cdots\!-P)}_{\text{GoF}_1(L)} \quad
\underbrace{(I-P\!-\!\cdots\!-P)}_{\text{GoF}_2(L)} \quad
\cdots \quad
\underbrace{(I-P\!-\!\cdots\!-P)}_{\text{GoF}_k(L)},$}
\end{equation}
where $\text{GoF}_k(L)$ denotes the $k^{th}$ group containing one intra-coded (I) frame followed by $L-1$ predictively-coded (P) frames. The model is trained end-to-end with a composite loss function:
\begin{equation}
L_{\text{total}} = L_{\text{render}} + \lambda_{\text{size}} L_{\text{rate}},
\end{equation}
where $L_{\text{render}}$ replicates the original 3D Gaussian Splatting objective, and $\lambda_{\text{size}}$ controls the rate-distortion trade-off.

\paragraph{\textbf{Encoding and Decoding Process.}}
The encoder extracts sparse motion from control points and compresses motion tensors $\boldsymbol{\hat{y}^m_t}$ and hyperpriors $\boldsymbol{\hat{z}^m_t}$ via arithmetic coding. During decoding, the system first reconstructs $\boldsymbol{\hat{z}^m_t}$, then combines it with spatial-temporal priors to estimate distribution parameters ($\boldsymbol{\sigma^m_t}$, $\boldsymbol{\mu^m_t}$) for decoding $\boldsymbol{\hat{y}^m_t}$. These decoded motion features enable subsequent motion compensation.

\section{Experiments}
We rigorously evaluate our D-FCGS model by addressing three critical research questions:
\begin{itemize}
\item \textbf{Generalization and Effectiveness:} Can D-FCGS generalize across new scenes from widely used datasets while achieving competitive rate-distortion performance compared to optimization-based methods? (\cref{sec:Q1})
\item \textbf{Robustness and Stability:} How does D-FCGS perform under diverse and high-dynamic scenes, and is the system stable under varying hyperparameters? (\cref{sec:Q2})
\item \textbf{Module Efficacy:} Do individual modules (e.g. control points) contribute meaningfully? (\cref{sec:Q3})
\end{itemize}

\subsection{Experimental Setup}
\label{sec:exp}

\paragraph{\textbf{Datasets and Implementation Details.}} We derive sequential Gaussian frames from six multi-view video datasets: (1) \textit{N3V} \cite{li2022neural} (2) \textit{MeetRoom} \cite{li2022streaming} (3) \textit{WideRange4D} \cite{yang2025widerange4d} (4) \textit{Google Immersive} \cite{broxton2020immersive} (5) \textit{Self-Cap} \cite{xu2024longvolcap} (6) \textit{VRU} \cite{wu2025swift4d}. For training, we use 3 scenes from \textit{MeetRoom} and 28 scenes from \textit{WideRange4D}. For evaluation, we reserve the "discussion" scene from \textit{MeetRoom} for in-domain testing and six scenes from \textit{N3V} for out-of-domain benchmarking. Additional robustness tests are performed on diverse and high-dynamic scenes (see \cref{sec:Q2}). All experiments are conducted on NVIDIA RTX 4090 GPU. Key hyperparameters include: (1) the downscale factor $M$ = 70, (2) $K$ = 30 for KNN, (3) quantization step $q'$ = 1, (4) GoF size $L=10$ during inference, and (4) $\lambda_{\text{size}} $ = 1e-3. Further details of datasets and implementation are provided in the supplementary material.

\begin{figure*}[t]
  \centering
   \includegraphics[width=1\linewidth]{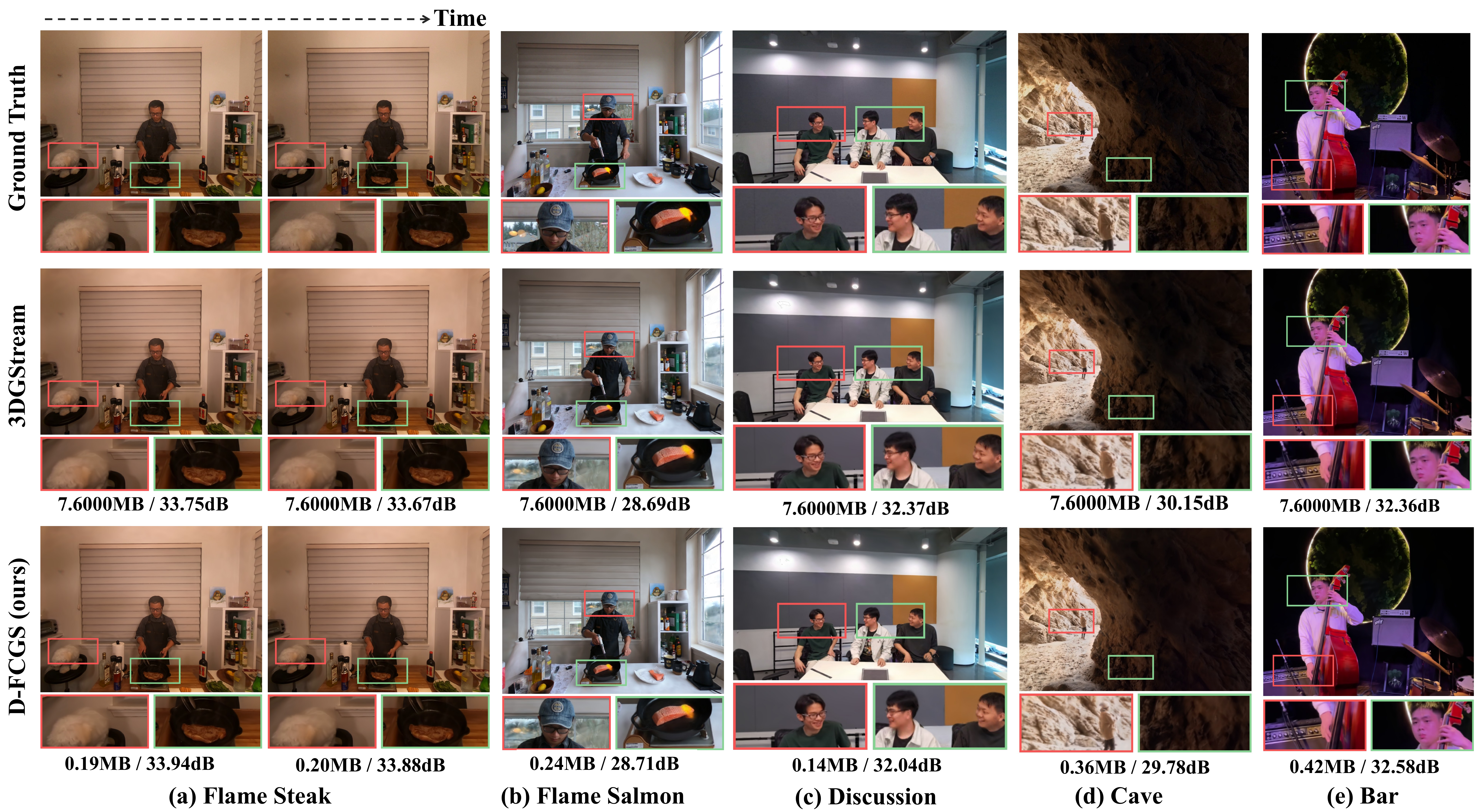}
\caption{\textbf{Qualitative comparison of rendering results.} Visual comparisons between (1) Ground Truth, (2) 3DGStream, and (3) our D-FCGS. D-FCGS significantly reduces storage size while maintaining comparable high fidelity to 3DGStream.}
   \label{fig:showcase}
\end{figure*}

\paragraph{\textbf{Evaluation Metrics.}} We evaluate D-FCGS using three kinds of metrics: (1) PSNR and SSIM \cite{wang2004image} for reconstruction quality, (2) compressed size (MB/frame) for rate efficiency, and (3) rendering speed (FPS) plus encoding/decoding time (sec) for computational performance.

\begin{table}[tb]
    \centering
    \small
    \setlength{\tabcolsep}{2pt}
        \begin{tabular}{c|cccc|c}
            \toprule
            Method & \begin{tabular}[c]{@{}c@{}}PSNR\\ (dB) ↑\end{tabular} & SSIM ↑ &  \begin{tabular}[c]{@{}c@{}}Size\\(MB) ↓\end{tabular} & \begin{tabular}[c]{@{}c@{}}Render\\ (FPS) ↑\end{tabular}   & \begin{tabular}[c]{@{}c@{}}Feedforward\\ Compression\end{tabular} \\ 
            \midrule
            K-Planes & 31.63& 0.920 & 1.0 & 0.15 &\ding{55} \\ 
            HyperReel & 31.10 & 0.931 & 1.7 & 16.7 &\ding{55} \\ 
            NeRFPlayer & 30.69 & 0.931 & 18.4& 0.05 & \ding{55} \\ 
            StreamRF & 30.61 & 0.930 & 7.6 & 8.3 & \ding{55} \\ 
            ReRF & 29.71 & 0.918 & 0.77 & 2.0 &\ding{55} \\ 
            TeTriRF & 30.65 & 0.931 & 0.76 & 2.7 & \ding{55} \\ 
            D-3DG & 30.67 & 0.931 & 9.2 & \textbf{460} & \ding{55} \\ 
            3DGStream* & \textbf{32.20} & \textbf{0.953} & 7.75 & 215 & \ding{55} \\ 
            HiCoM & 31.17 & - & 0.70 & \underline{274} & \ding{55} \\ 
            QUEEN & \underline{32.19} & 0.946 & 0.75 & 248 & \ding{55} \\ 
            4DGC & 31.58 & 0.943 & \underline{0.49} & 168 & \ding{55} \\              
            \textbf{D-FCGS (ours)} &  31.91 & \underline{0.952} & \textbf{0.46} & 215 &  \ding{51}\\ \bottomrule
        \end{tabular}
    \label{tab:n3v_result}
    \caption{
    Quantitative results on N3V \cite{li2022neural} dataset, averaged over 300 frames across six scenes. * denotes results reproduced by our implementation. \textbf{Bold} and \underline{underlined} values indicate the best and second-best performance, respectively. Detailed per-scene results are reported in supplementary material.
    }
\end{table}

\begin{table}[t]
    \centering
    \small
    \setlength{\tabcolsep}{7pt}
    \begin{tabular}{c|cccc}
        \toprule
        Method & \begin{tabular}[c]{@{}c@{}}PSNR\\ (dB) ↑\end{tabular} & SSIM ↑ &  \begin{tabular}[c]{@{}c@{}}Size\\(MB) ↓\end{tabular} & \begin{tabular}[c]{@{}c@{}}Render\\ (FPS) ↑\end{tabular}  \\ 
        \midrule
        StreamRF & 26.71 & 0.913 & 8.23 & 10  \\ 
        ReRF & 26.43 & 0.911 & 0.63 & 2.9 \\ 
        TeTriRF & 27.37 & 0.917 & 0.61 & 3.8  \\ 
        3DGStream* & \textbf{31.74} & \textbf{0.957} & 7.66 & \textbf{288} \\ 
        HiCoM & 29.61 & - & \underline{0.40}& \underline{284} \\ 
        4DGC & 28.08 & 0.922 & 0.42 & 213 \\ 
        \textbf{D-FCGS (ours)} & \underline{30.97} & \underline{0.950} & \textbf{0.38} & \textbf{288} \\ 
        \bottomrule
    \end{tabular}
    \label{tab:meetroom_result}
    \caption{
    Quantitative results on MeetRoom \cite{li2022streaming} dataset, averaged over 300 frames.
    }
\end{table}

\begin{table*}[t]
  \small
  \centering
  \label{tab:robustness}
  \begin{tabular}{c|ccc|ccc|ccc}
    \toprule
      & \multicolumn{3}{c|}{\textbf{Google Immersive}}
 & \multicolumn{3}{c|}{\textbf{Self-Cap}}  & \multicolumn{3}{c}{\textbf{VRU}}  \\
Method & PSNR(dB)↑ & SSIM↑ & Size(MB)↓ & PSNR(dB)↑ & SSIM↑ & Size(MB)↓ & PSNR(dB)↑ & SSIM↑ & Size(MB)↓  \\
    \midrule
     3DGStream* &25.84& 0.872& 7.6 & 26.43 & 0.858 & 7.6 & 26.18 & 0.892 & 7.6 \\
     D-FCGS(ours) & 25.58& 0.879 &0.34& 26.00 & 0.854 & 0.90 & 25.13 & 0.876 & 0.43\\
    \bottomrule
\end{tabular}
  \caption{Robustness test on diverse scenes from extensive datasets. Averaged PSNR, SSIM and P-frame size are reported. Per-scene results are provided in the appendix.}
\end{table*}

\begin{table}[h]
  \centering
  \label{tab:hyperparameters_robustness}
   \small
   \setlength{\tabcolsep}{2pt}
   
  \begin{tabular}{c|ccccccc}
    \toprule
    $K$ & 30 & 60 & 120 & 30 & 30 & 30 & 30 \\

    $M$   & 70 & 70 & 70 & 140 & 280 & 70 & 70 \\

    $L$   & 300 & 300 & 300 & 300 & 300 & 32 & 4\\
    \midrule
    Size↓ & 8.9 & 9.0 & 9.0 & 4.0 & 2.0 & 9.0 & 9.2\\
     PSNR↑  & 33.596 & 33.596 & 33.596 & 33.599 & 33.599 & 33.604 & 33.598\\
    \bottomrule
  \end{tabular}
    \caption{Robustness test on hyperparameter settings ($K$ for KNN, downscale factor $M$, and GoF size $L$). Averaged P-frame size ($\times$0.1 MB) and PSNR (dB) are reported on "cut\_beef" scene.}
\end{table}

\begin{table}[t]
  \centering
  \label{tab:time}
  \small
  \setlength{\tabcolsep}{4pt}
  \begin{tabular}{c|ccc}
    \toprule
    Method&Encoding(sec) & Decoding(sec)  & Total(sec) \\
    \midrule
     proposed &0.61& 0.72& 1.33 \\
     w/o control points &1.33& 2.88& 4.21 \\
    \bottomrule
    \end{tabular}
  \caption{Average encoding and decoding time for P-frames.}
\end{table}

\subsection{Benchmark Comparison}
\label{sec:Q1}

\paragraph{\textbf{Benchmark Methods.}} As the first feedforward inter-frame codec for Gaussian point clouds (to our knowledge), D-FCGS faces unique evaluation challenges due to the absence of directly comparable optimization-free approaches. Our analysis therefore encompasses two categories of benchmark methods in optimization-based dynamic scene reconstruction: (1) NeRF-based techniques including K-Planes \cite{fridovich2023k}, HyperReel \cite{attal2023hyperreel}, NeRFPlayer \cite{song2023nerfplayer}, StreamRF \cite{li2022streaming}, ReRF \cite{wang2023neural}, and TeTriRF \cite{wu2024tetrirf}; and (2) \textit{Per-Frame Gaussian} methods including D-3DG \cite{luiten2024dynamic}, 3DGStream \cite{sun20243dgstream}, HiCoM \cite{gao2024hicom}, QUEEN \cite{girish2024queen}, and 4DGC \cite{hu20254dgcrateaware4dgaussian}. While this comparison inherently favors optimization-based approaches  that benefit from scene-specific tuning, it provides critical insights into D-FCGS's performance relative to current paradigms in dynamic scene compression. 

\begin{figure*}[t]
  \centering
   \includegraphics[width=1\linewidth]{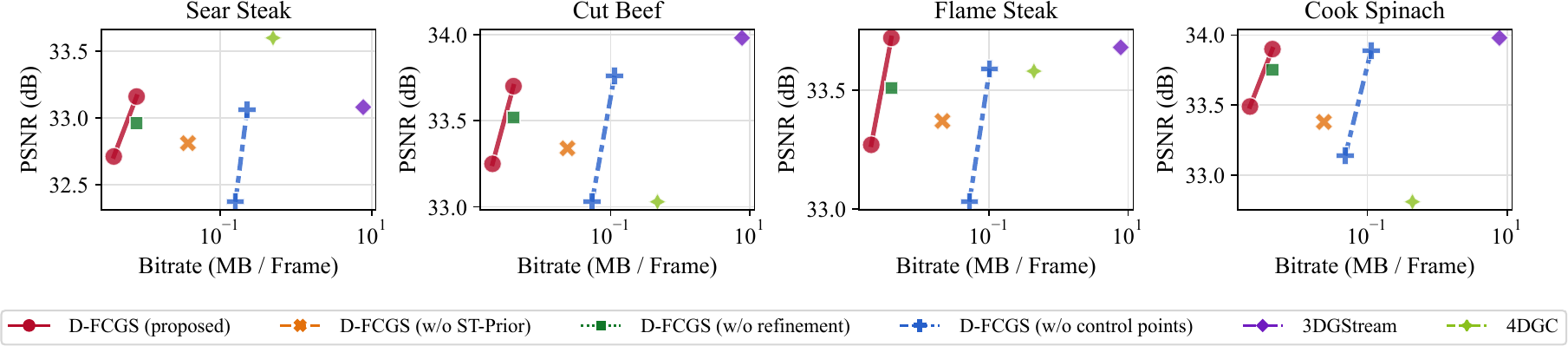}
   \caption{Rate-Distortion comparison of the proposed method and ablations.}
   \label{fig:aba_rd}
\end{figure*}


\paragraph{\textbf{Quantitative Results.}}

 Our compression results demonstrate significant improvements over existing methods. As shown in \cref{tab:n3v_result} and \cref{tab:meetroom_result}, D-FCGS achieves remarkable average sizes of \textbf{0.46MB} (\textit{N3V}) and \textbf{0.38MB} (\textit{MeetRoom}) per frame, namely a \textbf{17×} reduction compared to 3DGStream's 7.75MB and 7.66MB. Note that our codec is designed for inter-frame compression, and our P-frame compression ratio exceeds 32× in most cases (\cref{fig:teaser}). While I-frames could be further compressed using existing static methods (e.g., FCGS \cite{chen2024fast}), we maintain their original size for fair comparison. Additionally, the entire encoding/decoding pipeline operates efficiently, with both processes completing in under 1 second (\cref{tab:time}).

\paragraph{\textbf{Qualitative Results.}} We visualize the qualitative comparisons with 3DGStream \cite{sun20243dgstream} in \cref{fig:showcase}, presenting results on "flame steak", "flame salmon" (\textit{N3V} dataset) and "discussion" (\textit{MeetRoom} dataset). From scene details, we can tell that D-FCGS achieves comparable fidelity to 3DGStream, even better in some cases.

\subsection{Robustness Test}
\label{sec:Q2}
\paragraph{Robustness to Diverse Scenes.}
We evaluate D-FCGS on three additional datasets: \textit{Google Immersive}, \textit{Self-Cap}, and \textit{VRU}, covering scenarios such as indoor painting, cave exploration, and basketball games. As summarized in \cref{tab:robustness}, for Gaussian sequences reconstructed by 3DGStream, our approach achieves unprecedented P-frame compression while maintaining near-identical SSIM and small PSNR drop ($<$ 0.5 dB for \textit{Google Immersive} and \textit{Self-Cap}). While PSNR gaps exist for \textit{VRU} (full of blurry motions), our method offers a more practical and efficient pipeline for real-world applications, sacrificing acceptable visual fidelity for great improvements in bitrates. \cref{fig:showcase} highlights visual results on the outdoor "cave" (\textit{Google Immersive}) and high-dynamic "bar" (\textit{Self-Cap}). 


\paragraph{Robustness to Hyperparameter Settings.} Here, we apply varied parameter configurations (KNN neighborhood size $K$, downscale factor $M$, and GoF size $L$) to the "cut beef" scene (\textit{N3V}). As presented in \cref{tab:hyperparameters_robustness}, D-FCGS maintains stable rendering quality across most parameter combinations, showing strong parametric robustness. Notably, increasing the downscale factor $M$ effectively reduces control point counts, leading to a linear decrease in P-frame compression size, which aligns with expectations.

\subsection{Ablation Study}
\label{sec:Q3}

In this section, we evaluate key components of D-FCGS through systematic ablation studies.

\paragraph{\textbf{Effect of Control Points.}} The sparse motion representation via control points forms the foundation of our efficient compression pipeline. Removing control points and predicting motions for all Gaussians increases storage substantially (\cref{fig:aba_rd}) and slows encoding/decoding by 3.2× (\cref{tab:time}). This validates our sparse motion representation's efficiency in rate saving and computational cost.

\paragraph{\textbf{Effect of Dual Prior-Aware Entropy Model.}} 
 Our entropy model employs a novel dual-prior architecture. While hyperpriors effectively capture global dependencies  \cite{balle2018variational}, our key innovation lies in the hash-grid-based spatial-temporal prior that models local correlations. Removing the spatial-temporal prior branch leads to noticeable R-D performance degradation (\cref{fig:aba_rd}), confirming its importance to optimal compression.

\paragraph{\textbf{Effect of Color Refinement.}} 

Our online color refinement module improves rendering quality (PSNR) by  0.1 $\sim$ 0.5 dB (\cref{fig:aba_rd}), while requiring no additional storage and  negligible decoding overhead. Per-scene results on \textit{N3V} dataset are shown in the supplementary.


\section{Conclusion}

In this paper, we propose \textbf{F}eedforward \textbf{C}ompression of \textbf{D}ynamic \textbf{G}aussian \textbf{S}platting (D-FCGS), a novel feedforward framework for zero-shot dynamic Gaussian sequence compression. 
Our contributions are threefold. 
First, we adopt the I-P coding profile for standard GS compression and introduce sparse inter-frame motion extraction via control points. 
Second, we present an end-to-end motion compression framework with a dual prior-aware entropy model, fully leveraging hyperpriors and spatial-temporal context to improve rate estimation. 
Third, control-point-guided motion compensation is combined with a color refinement network to guarantee high-fidelity and view-consistent reconstruction. 
Experimental results show that \emph{D-FCGS} achieves superior compression efficiency (over 17× compression) on two benchmark datasets (MeetRoom and N3V) and remarkable robustness across diverse scenes from extensive datasets, significantly enhancing transmission and storage efficiency for free-viewpoint video applications.

\section*{Acknowledgments}
This work was partly supported by the NSFC62431015, Science and Technology Commission of Shanghai Municipality No.24511106200, the Fundamental Research Funds for the Central Universities, Shanghai Key Laboratory of Digital Media Processing and Transmission under Grant 22DZ2229005, 111 project BP0719010, Okawa Research Grant and Explore-X Research Fund.

\bibliography{aaai2026}

@article{yang2025hybridgs,
  title={HybridGS: High-Efficiency Gaussian Splatting Data Compression using Dual-Channel Sparse Representation and Point Cloud Encoder},
  author={Yang, Qi and Yang, Le and Van Der Auwera, Geert and Li, Zhu},
  journal={arXiv preprint arXiv:2505.01938},
  year={2025}
}

@inproceedings{huang2025hierarchical,
  title={A hierarchical compression technique for 3d gaussian splatting compression},
  author={Huang, He and Huang, Wenjie and Yang, Qi and Xu, Yiling and Li, Zhu},
  booktitle={ICASSP 2025-2025 IEEE International Conference on Acoustics, Speech and Signal Processing (ICASSP)},
  pages={1--5},
  year={2025},
  organization={IEEE}
}

@article{kerbl20233d,
  title={3d gaussian splatting for real-time radiance field rendering.},
  author={Kerbl, Bernhard and Kopanas, Georgios and Leimk{\"u}hler, Thomas and Drettakis, George},
  journal={ACM Trans. Graph.},
  volume={42},
  number={4},
  pages={139--1},
  year={2023}
}

@inproceedings{lee2024compact,
  title={Compact 3d gaussian representation for radiance field},
  author={Lee, Joo Chan and Rho, Daniel and Sun, Xiangyu and Ko, Jong Hwan and Park, Eunbyung},
  booktitle={Proceedings of the IEEE/CVF Conference on Computer Vision and Pattern Recognition},
  pages={21719--21728},
  year={2024}
}

@article{fan2024lightgaussian,
  title={Lightgaussian: Unbounded 3d gaussian compression with 15x reduction and 200+ fps},
  author={Fan, Zhiwen and Wang, Kevin and Wen, Kairun and Zhu, Zehao and Xu, Dejia and Wang, Zhangyang and others},
  journal={Advances in neural information processing systems},
  volume={37},
  pages={140138--140158},
  year={2024}
}

@inproceedings{lu2024scaffold,
  title={Scaffold-gs: Structured 3d gaussians for view-adaptive rendering},
  author={Lu, Tao and Yu, Mulin and Xu, Linning and Xiangli, Yuanbo and Wang, Limin and Lin, Dahua and Dai, Bo},
  booktitle={Proceedings of the IEEE/CVF Conference on Computer Vision and Pattern Recognition},
  pages={20654--20664},
  year={2024}
}

@inproceedings{chen2024hac,
  title={Hac: Hash-grid assisted context for 3d gaussian splatting compression},
  author={Chen, Yihang and Wu, Qianyi and Lin, Weiyao and Harandi, Mehrtash and Cai, Jianfei},
  booktitle={European Conference on Computer Vision},
  pages={422--438},
  year={2024},
  organization={Springer}
}

@article{chen2025hac++,
  title={HAC++: Towards 100X Compression of 3D Gaussian Splatting},
  author={Chen, Yihang and Wu, Qianyi and Lin, Weiyao and Harandi, Mehrtash and Cai, Jianfei},
  journal={arXiv preprint arXiv:2501.12255},
  year={2025}
}

@inproceedings{niedermayr2024compressed,
  title={Compressed 3d gaussian splatting for accelerated novel view synthesis},
  author={Niedermayr, Simon and Stumpfegger, Josef and Westermann, R{\"u}diger},
  booktitle={Proceedings of the IEEE/CVF Conference on Computer Vision and Pattern Recognition},
  pages={10349--10358},
  year={2024}
}

@article{navaneet2023compact3d,
  title={Compact3d: Compressing gaussian splat radiance field models with vector quantization},
  author={Navaneet, K and Meibodi, Kossar Pourahmadi and Koohpayegani, Soroush Abbasi and Pirsiavash, Hamed},
  journal={arXiv preprint arXiv:2311.18159},
  volume={4},
  year={2023}
}

@inproceedings{girish2024eagles,
  title={Eagles: Efficient accelerated 3d gaussians with lightweight encodings},
  author={Girish, Sharath and Gupta, Kamal and Shrivastava, Abhinav},
  booktitle={European Conference on Computer Vision},
  pages={54--71},
  year={2024},
  organization={Springer}
}

@article{ali2024trimming,
  title={Trimming the fat: Efficient compression of 3d gaussian splats through pruning},
  author={Ali, Muhammad Salman and Qamar, Maryam and Bae, Sung-Ho and Tartaglione, Enzo},
  journal={arXiv preprint arXiv:2406.18214},
  year={2024}
}

@article{zhan2025cat,
  title={CAT-3DGS: A Context-Adaptive Triplane Approach to Rate-Distortion-Optimized 3DGS Compression},
  author={Zhan, Yu-Ting and Ho, Cheng-Yuan and Yang, Hebi and Chen, Yi-Hsin and Chiang, Jui Chiu and Liu, Yu-Lun and Peng, Wen-Hsiao},
  journal={arXiv preprint arXiv:2503.00357},
  year={2025}
}

@article{balle2018variational,
  title={Variational image compression with a scale hyperprior},
  author={Ball{\'e}, Johannes and Minnen, David and Singh, Saurabh and Hwang, Sung Jin and Johnston, Nick},
  journal={arXiv preprint arXiv:1802.01436},
  year={2018}
}

@article{balle2016end,
  title={End-to-end optimized image compression},
  author={Ball{\'e}, Johannes and Laparra, Valero and Simoncelli, Eero P},
  journal={arXiv preprint arXiv:1611.01704},
  year={2016}
}

@article{lee2025compression,
  title={Compression of 3D Gaussian Splatting with Optimized Feature Planes and Standard Video Codecs},
  author={Lee, Soonbin and Shu, Fangwen and Sanchez, Yago and Schierl, Thomas and Hellge, Cornelius},
  journal={arXiv preprint arXiv:2501.03399},
  year={2025}
}

@inproceedings{morgenstern2024compact,
  title={Compact 3d scene representation via self-organizing gaussian grids},
  author={Morgenstern, Wieland and Barthel, Florian and Hilsmann, Anna and Eisert, Peter},
  booktitle={European Conference on Computer Vision},
  pages={18--34},
  year={2024},
  organization={Springer}
}

@inproceedings{wang2024end,
  title={End-to-end rate-distortion optimized 3d gaussian representation},
  author={Wang, Henan and Zhu, Hanxin and He, Tianyu and Feng, Runsen and Deng, Jiajun and Bian, Jiang and Chen, Zhibo},
  booktitle={European Conference on Computer Vision},
  pages={76--92},
  year={2024},
  organization={Springer}
}

@inproceedings{cheng2020learned,
  title={Learned image compression with discretized gaussian mixture likelihoods and attention modules},
  author={Cheng, Zhengxue and Sun, Heming and Takeuchi, Masaru and Katto, Jiro},
  booktitle={Proceedings of the IEEE/CVF conference on computer vision and pattern recognition},
  pages={7939--7948},
  year={2020}
}

@article{chen2024fast,
  title={Fast feedforward 3d gaussian splatting compression},
  author={Chen, Yihang and Wu, Qianyi and Li, Mengyao and Lin, Weiyao and Harandi, Mehrtash and Cai, Jianfei},
  journal={arXiv preprint arXiv:2410.08017},
  year={2024}
}

@inproceedings{yang2024deformable,
  title={Deformable 3d gaussians for high-fidelity monocular dynamic scene reconstruction},
  author={Yang, Ziyi and Gao, Xinyu and Zhou, Wen and Jiao, Shaohui and Zhang, Yuqing and Jin, Xiaogang},
  booktitle={Proceedings of the IEEE/CVF conference on computer vision and pattern recognition},
  pages={20331--20341},
  year={2024}
}

@inproceedings{wu20244d,
  title={4d gaussian splatting for real-time dynamic scene rendering},
  author={Wu, Guanjun and Yi, Taoran and Fang, Jiemin and Xie, Lingxi and Zhang, Xiaopeng and Wei, Wei and Liu, Wenyu and Tian, Qi and Wang, Xinggang},
  booktitle={Proceedings of the IEEE/CVF conference on computer vision and pattern recognition},
  pages={20310--20320},
  year={2024}
}

@inproceedings{li2024spacetime,
  title={Spacetime gaussian feature splatting for real-time dynamic view synthesis},
  author={Li, Zhan and Chen, Zhang and Li, Zhong and Xu, Yi},
  booktitle={Proceedings of the IEEE/CVF Conference on Computer Vision and Pattern Recognition},
  pages={8508--8520},
  year={2024}
}

@inproceedings{duan20244d,
  title={4d-rotor gaussian splatting: towards efficient novel view synthesis for dynamic scenes},
  author={Duan, Yuanxing and Wei, Fangyin and Dai, Qiyu and He, Yuhang and Chen, Wenzheng and Chen, Baoquan},
  booktitle={ACM SIGGRAPH 2024 Conference Papers},
  pages={1--11},
  year={2024}
}

@inproceedings{sun20243dgstream,
  title={3dgstream: On-the-fly training of 3d gaussians for efficient streaming of photo-realistic free-viewpoint videos},
  author={Sun, Jiakai and Jiao, Han and Li, Guangyuan and Zhang, Zhanjie and Zhao, Lei and Xing, Wei},
  booktitle={Proceedings of the IEEE/CVF Conference on Computer Vision and Pattern Recognition},
  pages={20675--20685},
  year={2024}
}

@inproceedings{lin2024gaussian,
  title={Gaussian-flow: 4d reconstruction with dynamic 3d gaussian particle},
  author={Lin, Youtian and Dai, Zuozhuo and Zhu, Siyu and Yao, Yao},
  booktitle={Proceedings of the IEEE/CVF Conference on Computer Vision and Pattern Recognition},
  pages={21136--21145},
  year={2024}
}

@inproceedings{yan2025instant,
  title={Instant gaussian stream: Fast and generalizable streaming of dynamic scene reconstruction via gaussian splatting},
  author={Yan, Jinbo and Peng, Rui and Wang, Zhiyan and Tang, Luyang and Yang, Jiayu and Liang, Jie and Wu, Jiahao and Wang, Ronggang},
  booktitle={Proceedings of the Computer Vision and Pattern Recognition Conference},
  pages={16520--16531},
  year={2025}
}

@article{gao2024hicom,
  title={Hicom: Hierarchical coherent motion for dynamic streamable scenes with 3d gaussian splatting},
  author={Gao, Qiankun and Meng, Jiarui and Wen, Chengxiang and Chen, Jie and Zhang, Jian},
  journal={Advances in Neural Information Processing Systems},
  volume={37},
  pages={80609--80633},
  year={2024}
}

@article{he2024s4d,
  title={S4d: Streaming 4d real-world reconstruction with gaussians and 3d control points},
  author={He, Bing and Chen, Yunuo and Lu, Guo and Wang, Qi and Gu, Qunshan and Xie, Rong and Song, Li and Zhang, Wenjun},
  journal={arXiv preprint arXiv:2408.13036},
  year={2024}
}

@article{lee2024fully,
  title={Fully explicit dynamic gaussian splatting},
  author={Lee, Junoh and Won, ChangYeon and Jung, Hyunjun and Bae, Inhwan and Jeon, Hae-Gon},
  journal={Advances in Neural Information Processing Systems},
  volume={37},
  pages={5384--5409},
  year={2024}
}

@article{yang2023real,
  title={Real-time photorealistic dynamic scene representation and rendering with 4d gaussian splatting},
  author={Yang, Zeyu and Yang, Hongye and Pan, Zijie and Zhang, Li},
  journal={arXiv preprint arXiv:2310.10642},
  year={2023}
}

@article{girish2024queen,
  title={QUEEN: QUantized Efficient ENcoding of Dynamic Gaussians for Streaming Free-viewpoint Videos},
  author={Girish, Sharath and Li, Tianye and Mazumdar, Amrita and Shrivastava, Abhinav and De Mello, Shalini and others},
  journal={Advances in Neural Information Processing Systems},
  volume={37},
  pages={43435--43467},
  year={2024}
}

@article{papantonakis2024reducing,
  title={Reducing the memory footprint of 3d gaussian splatting},
  author={Papantonakis, Panagiotis and Kopanas, Georgios and Kerbl, Bernhard and Lanvin, Alexandre and Drettakis, George},
  journal={Proceedings of the ACM on Computer Graphics and Interactive Techniques},
  volume={7},
  number={1},
  pages={1--17},
  year={2024},
  publisher={ACM New York, NY, USA}
}

@inproceedings{huang2024sc,
  title={Sc-gs: Sparse-controlled gaussian splatting for editable dynamic scenes},
  author={Huang, Yi-Hua and Sun, Yang-Tian and Yang, Ziyi and Lyu, Xiaoyang and Cao, Yan-Pei and Qi, Xiaojuan},
  booktitle={Proceedings of the IEEE/CVF conference on computer vision and pattern recognition},
  pages={4220--4230},
  year={2024}
}

@article{wang2024v,
  title={V\^{} 3: Viewing Volumetric Videos on Mobiles via Streamable 2D Dynamic Gaussians},
  author={Wang, Penghao and Zhang, Zhirui and Wang, Liao and Yao, Kaixin and Xie, Siyuan and Yu, Jingyi and Wu, Minye and Xu, Lan},
  journal={ACM Transactions on Graphics (TOG)},
  volume={43},
  number={6},
  pages={1--13},
  year={2024},
  publisher={ACM New York, NY, USA}
}

@inproceedings{luiten2024dynamic,
  title={Dynamic 3d gaussians: Tracking by persistent dynamic view synthesis},
  author={Luiten, Jonathon and Kopanas, Georgios and Leibe, Bastian and Ramanan, Deva},
  booktitle={2024 International Conference on 3D Vision (3DV)},
  pages={800--809},
  year={2024},
  organization={IEEE}
}

@article{wu2025swift4d,
  title={Swift4D: Adaptive divide-and-conquer Gaussian Splatting for compact and efficient reconstruction of dynamic scene},
  author={Wu, Jiahao and Peng, Rui and Wang, Zhiyan and Xiao, Lu and Tang, Luyang and Yan, Jinbo and Xiong, Kaiqiang and Wang, Ronggang},
  journal={arXiv preprint arXiv:2503.12307},
  year={2025}
}

@misc{hu20254dgcrateaware4dgaussian,
      title={4DGC: Rate-Aware 4D Gaussian Compression for Efficient Streamable Free-Viewpoint Video}, 
      author={Qiang Hu and Zihan Zheng and Houqiang Zhong and Sihua Fu and Li Song and XiaoyunZhang and Guangtao Zhai and Yanfeng Wang},
      year={2025},
      eprint={2503.18421},
      archivePrefix={arXiv},
      primaryClass={cs.CV},
      url={https://arxiv.org/abs/2503.18421}, 
}

@article{yang2025widerange4d,
  title={WideRange4D: Enabling High-Quality 4D Reconstruction with Wide-Range Movements and Scenes},
  author={Yang, Ling and Zhu, Kaixin and Tian, Juanxi and Zeng, Bohan and Lin, Mingbao and Pei, Hongjuan and Zhang, Wentao and Yan, Shuicheng},
  journal={arXiv preprint arXiv:2503.13435},
  year={2025}
}

@inproceedings{li2022neural,
  title={Neural 3d video synthesis from multi-view video},
  author={Li, Tianye and Slavcheva, Mira and Zollhoefer, Michael and Green, Simon and Lassner, Christoph and Kim, Changil and Schmidt, Tanner and Lovegrove, Steven and Goesele, Michael and Newcombe, Richard and others},
  booktitle={Proceedings of the IEEE/CVF conference on computer vision and pattern recognition},
  pages={5521--5531},
  year={2022}
}

@article{broxton2020immersive,
  title={Immersive light field video with a layered mesh representation},
  author={Broxton, Michael and Flynn, John and Overbeck, Ryan and Erickson, Daniel and Hedman, Peter and Duvall, Matthew and Dourgarian, Jason and Busch, Jay and Whalen, Matt and Debevec, Paul},
  journal={ACM Transactions on Graphics (TOG)},
  volume={39},
  number={4},
  pages={86--1},
  year={2020},
  publisher={ACM New York, NY, USA}
}

@inproceedings{fridovich2023k,
  title={K-planes: Explicit radiance fields in space, time, and appearance},
  author={Fridovich-Keil, Sara and Meanti, Giacomo and Warburg, Frederik Rahb{\ae}k and Recht, Benjamin and Kanazawa, Angjoo},
  booktitle={Proceedings of the IEEE/CVF Conference on Computer Vision and Pattern Recognition},
  pages={12479--12488},
  year={2023}
}

@inproceedings{attal2023hyperreel,
  title={HyperReel: High-fidelity 6-DoF video with ray-conditioned sampling},
  author={Attal, Benjamin and Huang, Jia-Bin and Richardt, Christian and Zollhoefer, Michael and Kopf, Johannes and O’Toole, Matthew and Kim, Changil},
  booktitle={Proceedings of the IEEE/CVF Conference on Computer Vision and Pattern Recognition},
  pages={16610--16620},
  year={2023}
}

@article{song2023nerfplayer,
  title={Nerfplayer: A streamable dynamic scene representation with decomposed neural radiance fields},
  author={Song, Liangchen and Chen, Anpei and Li, Zhong and Chen, Zhang and Chen, Lele and Yuan, Junsong and Xu, Yi and Geiger, Andreas},
  journal={IEEE Transactions on Visualization and Computer Graphics},
  volume={29},
  number={5},
  pages={2732--2742},
  year={2023},
  publisher={IEEE}
}

@article{li2022streaming,
  title={Streaming radiance fields for 3d video synthesis},
  author={Li, Lingzhi and Shen, Zhen and Wang, Zhongshu and Shen, Li and Tan, Ping},
  journal={Advances in Neural Information Processing Systems},
  volume={35},
  pages={13485--13498},
  year={2022}
}

@inproceedings{wang2023neural,
  title={Neural residual radiance fields for streamably free-viewpoint videos},
  author={Wang, Liao and Hu, Qiang and He, Qihan and Wang, Ziyu and Yu, Jingyi and Tuytelaars, Tinne and Xu, Lan and Wu, Minye},
  booktitle={Proceedings of the IEEE/CVF Conference on Computer Vision and Pattern Recognition},
  pages={76--87},
  year={2023}
}

@inproceedings{wu2024tetrirf,
  title={Tetrirf: Temporal tri-plane radiance fields for efficient free-viewpoint video},
  author={Wu, Minye and Wang, Zehao and Kouros, Georgios and Tuytelaars, Tinne},
  booktitle={Proceedings of the IEEE/CVF conference on computer vision and pattern recognition},
  pages={6487--6496},
  year={2024}
}

@article{shannon1948mathematical,
  title={A mathematical theory of communication},
  author={Shannon, Claude E},
  journal={The Bell system technical journal},
  volume={27},
  number={3},
  pages={379--423},
  year={1948},
  publisher={Nokia Bell Labs}
}

@article{mildenhall2021nerf,
  title={Nerf: Representing scenes as neural radiance fields for view synthesis},
  author={Mildenhall, Ben and Srinivasan, Pratul P and Tancik, Matthew and Barron, Jonathan T and Ramamoorthi, Ravi and Ng, Ren},
  journal={Communications of the ACM},
  volume={65},
  number={1},
  pages={99--106},
  year={2021},
  publisher={ACM New York, NY, USA}
}

@inproceedings{zheng2024hpc,
  title={HPC: Hierarchical Progressive Coding Framework for Volumetric Video},
  author={Zheng, Zihan and Zhong, Houqiang and Hu, Qiang and Zhang, Xiaoyun and Song, Li and Zhang, Ya and Wang, Yanfeng},
  booktitle={Proceedings of the 32nd ACM International Conference on Multimedia},
  pages={7937--7946},
  year={2024}
}

@inproceedings{zheng2024jointrf,
  title={JointRF: End-to-End Joint Optimization for Dynamic Neural Radiance Field Representation and Compression},
  author={Zheng, Zihan and Zhong, Houqiang and Hu, Qiang and Zhang, Xiaoyun and Song, Li and Zhang, Ya and Wang, Yanfeng},
  booktitle={2024 IEEE International Conference on Image Processing (ICIP)},
  pages={3292--3298},
  year={2024},
  organization={IEEE}
}

@article{wang2004image,
  title={Image quality assessment: from error visibility to structural similarity},
  author={Wang, Zhou and Bovik, Alan C and Sheikh, Hamid R and Simoncelli, Eero P},
  journal={IEEE transactions on image processing},
  volume={13},
  number={4},
  pages={600--612},
  year={2004},
  publisher={IEEE}
}

@Article{xu2024longvolcap,
  author  = {Xu, Zhen and Xu, Yinghao and Yu, Zhiyuan and Peng, Sida and Sun, Jiaming and Bao, Hujun and Zhou, Xiaowei},
  title   = {Representing Long Volumetric Video with Temporal Gaussian Hierarchy},
  journal = {ACM Transactions on Graphics},
  number  = {6},
  volume  = {43},
  month   = {November},
  year    = {2024},
  url     = {https://zju3dv.github.io/longvolcap}
}

\end{document}